\documentclass[letterpaper, 10pt, conference]{ieeeconf}      

\IEEEoverridecommandlockouts                              

\usepackage{graphics} 
\usepackage{epsfig} 
\usepackage{mathptmx} 
\usepackage{times} 
\usepackage{amsmath} 
\usepackage{amssymb} 
\usepackage{xcolor}
\usepackage{booktabs}
\usepackage{graphicx}
\usepackage{multirow}
\usepackage{listings}
\usepackage{url}
\usepackage[utf8]{inputenc}
\usepackage{amsfonts}
\usepackage{algorithm}
\usepackage{algpseudocode}
\usepackage{hyperref}
\usepackage{float}

\title{\LARGE \bf
AlignBot: \underline{Align}ing VLM-powered Customized Task Planning with \\User Reminders Through Fine-Tuning for Household Ro\underline{bot}s
}

\author{Zhaxizhuoma$^{1\dagger}$, Pengan Chen$^{1,2\dagger}$, Ziniu Wu$^{1,3\dagger}$, Jiawei Sun$^{1}$, Dong Wang$^{1}$, \hspace{0.2cm}\\ Peng Zhou$^{2}$, Nieqing Cao$^{4}$, Yan Ding$^{1*}$, Bin Zhao$^{1,5}$, Xuelong Li$^{1,6}$
\thanks{$^\dagger$ The first three authors contribute equally.
$^*$ Correspondence to: Yan Ding \textless yding25@binghamton.edu\textgreater}\\
\thanks{$^1$~Shanghai Artificial Intelligence Laboratory, $^2$~The University of Hong Kong, $^3$~University of Bristol, $^4$~Xi'an Jiaotong-Liverpool University, $^5$~Northwestern Polytechnical University, $^6$~Institute of Artificial Intelligence, China Telecom Corp Ltd}
}

\definecolor{myred}{RGB}{255,0,0}
\definecolor{mygreen}{RGB}{0,176,80}
\definecolor{personalized_preferences_green}{RGB}{109,204,109}
\definecolor{personalized_preferences_blue}{RGB}{85,172,213}
\definecolor{personalized_preferences_yellow}{RGB}{255,223,96}

\begin{document}

\maketitle
\thispagestyle{empty}
\pagestyle{empty}

\begin{abstract}
This paper presents AlignBot, a novel framework designed to optimize VLM-powered customized task planning for household robots by effectively aligning with user reminders.
In domestic settings, aligning task planning with user reminders poses significant challenges due to the limited quantity, diversity, and multimodal nature of the reminders.
To address these challenges, AlignBot employs a fine-tuned LLaVA-7B model, functioning as an adapter for GPT-4o.
This adapter model internalizes diverse forms of user reminders-such as personalized preferences, corrective guidance, and contextual assistance-into structured \emph{instruction-formatted cues} that prompt GPT-4o in generating customized task plans.
Additionally, AlignBot integrates a dynamic retrieval mechanism that selects task-relevant historical successes as prompts for GPT-4o, further enhancing task planning accuracy.
To validate the effectiveness of AlignBot, experiments are conducted in real-world household environments, which are constructed within the laboratory to replicate typical household settings.
A multimodal dataset with over 1,500 entries derived from volunteer reminders is used for training and evaluation.
The results demonstrate that AlignBot significantly improves customized task planning, outperforming existing LLM- and VLM-powered planners by interpreting and aligning with user reminders, achieving 86.8\% success rate compared to the vanilla GPT-4o baseline at 21.6\%, reflecting a 65\% improvement and over four times greater effectiveness.
Supplementary materials are available at: \url{https://yding25.com/AlignBot/}
\end{abstract}

\section{Introduction}
Household robots hold the promise of becoming essential in human daily life, providing long-term customized service to families in their home environments~\cite{ding2023task,ding2023integrating,wu2023tidybot,zhang2023hierarchical}.
To operate effectively in different homes, robots need to adapt their foundational capabilities, such as task planning to individual users and their unique household contexts. 
To ensure user satisfaction, these robots need to interact naturally with users, collecting their reminders and continuously optimizing the customized task planning through iterative updates based on the collected data.

\begin{figure}[t]
    \centering\includegraphics[width=0.75\linewidth]{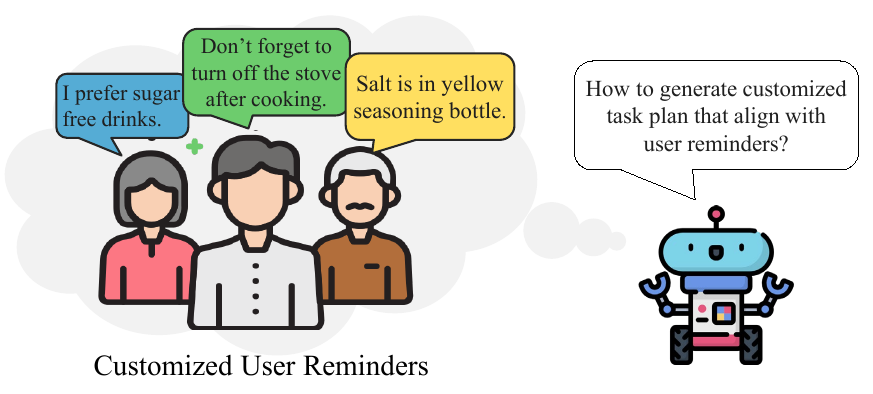}
    \vspace{-0.3em}
    \caption{The robot needs to align its task planning with customized user reminders-categorized into three types: \textcolor{personalized_preferences_blue}{personalized preferences}, \textcolor{personalized_preferences_green}{corrective guidance}, and \textcolor{personalized_preferences_yellow}{contextual assistance}-with each illustrated through examples in distinct colors.
    \vspace{-1.8em}}
    \label{fig:demo1}
\end{figure}

Vision-Language Models (VLMs) are emerging as a new paradigm in robotic task planning~\cite{driess2023palmeembodiedmultimodallanguage,vemprala2023chatgptroboticsdesignprinciples,huang2024rekepspatiotemporalreasoningrelational,Lin_2023,Wake_2023,ahn2024autortembodiedfoundationmodels,driess2023palm,zhu2023minigpt,zhou2024proreasonmultimodalproactivereasoning}.
Compared to Large Language Models (LLMs), VLMs can leverage both visual and language information, and obtain a more comprehensive understanding of the task context. 
However, \emph{aligning VLM-powered task planners for household robots with user reminders is challenging due to the limited quantity of reminders, their diversity in type, and their multimodal nature.}
In a home setting, user-robot interactions are relatively infrequent, resulting in a sparse reminder dataset, which makes it extremely challenging to fine-tune a VLM with a large number of parameters, such as GPT-4o~\cite{openai2024gpt4o}.
Moreover, user reminders are inherently diverse, including personalized preferences, corrective guidance, and contextual assistance~\cite{ren2023robotsaskhelpuncertainty,liu2023reflectsummarizingrobotexperiences,ding2023integrating,han2024llm,wu2023tidybot,shi2024yellrobotimprovingonthefly}, as illustrated in Fig.~\ref{fig:demo1}. 
This diversity necessitates a model with strong generalization capabilities to effectively process and learn from the varied reminders. 
Additionally, the multimodal nature of user reminders, which frequently involves both text and visual elements that are closely linked, increases the alignment complexity, posing challenges that single-modality models cannot fully address.

To the best of our knowledge, in the domain of robotics, few existing VLM-powered task planning approaches effectively align with customized user reminders.
Previous works on VLM grounding has primarily focused on aligning these models with the physical context of tasks by translating generated plans into executable actions and incorporating contextual information such as affordances or environmental feedback~\cite{zhi2024closed,hu2023look,liao2024can,mei2024gamevlm}.
Some works on LLMs in task planning have explored diverse alignment strategies~\cite{rafailov2024directpreferenceoptimizationlanguage,wu2023tidybot,Kim_2024,liu2024humanawarenessrobottask}.
The most closely related recent work is LLM-Personalized~\cite{han2024llm}, which connects an LLM planner with personalized user preferences, primarily within the domain of housekeeping, using reinforced self-training techniques.
However, this approach does not account for the broader range of user reminders, such as corrective guidance, leaving the challenge of aligning customized task planning with complex user reminders, particularly beyond housekeeping tasks, largely unaddressed.

To this end, we introduce \emph{AlignBot}, an advanced framework designed to effectively align diverse and multimodal user reminders with robotics customized task planning for everyday household activities. 
Within AlignBot, a fine-tuned LLaVA-7B model functions as an adapter for GPT-4o~\cite{liu2023improvedllava,openai2024gpt4o,hu2022lora,zhou2024empiricalstudyparameterefficientfinetuning,wang2025mosunleashingparameterefficiency,wang2024prolorapartialrotationempowers}, internalizing user reminders to generate \emph{instruction-formatted cues} that guide GPT-4o in task planning.
By ``freezing'' GPT-4o and fine-tuning only the much smaller LLaVA-7B model, we significantly reduce data requirements while leveraging LLaVA-7B's multimodal capabilities to handle multimodal reminders. 
Within our data collection pipeline, we address the diversity of user-robot interactions by generating standardized instruction-formatted cues that transform complex dialogue inputs into consistent prompts. 
These cues streamline the processing of diverse reminders and optimize GPT-4o's performance, given its strength in interpreting and executing structured prompts~\cite{ratnayake2023prompting}.
To address action omissions and sequence errors in long-term tasks, we also implement a \emph{dynamic retrieval mechanism} that selects relevant cases from historically successful task plans as prompts for GPT-4o, thereby improving task execution accuracy.

To empirically evaluate AlignBot, we establish a training setup that realistically simulates a household environment, consisting of three diverse scenarios and involving a collection of 81 objects.
Multiple volunteers participated by providing reminders during the robot's execution of 20 everyday tasks, resulting in a multimodal, structured dataset comprising over 1,500 entries for fine-tuning the LLaVA-7B model.
We assess the quality of the cues generated by LLaVA using real user ratings and measure the success rate of the task plans produced by GPT-4o.
The results demonstrate that AlignBot enhances task plan generation quality, outperforming the most advanced LLM- and VLM-powered planners, largely due to its superior capability to interpret and align with user reminders.

\section{Related Work}
\vspace{.5em}
\noindent
\textbf{VLM-Powered Robotic Task Planning:} VLMs have only recently emerged as a promising paradigm, rapidly gaining significant attention within the field~\cite{zhi2024closed,hu2023look,shirai2024vision,zhang2024dkprompt,huang2024rekepspatiotemporalreasoningrelational,Lin_2023,Wake_2023,ahn2024autortembodiedfoundationmodels,ren2024exploreconfidentefficientexploration}.
Unlike LLM-powered approaches~\cite{ding2023task,singh2022progpromptgeneratingsituatedrobot,joublin2023copalcorrectiveplanningrobot,touvron2023llama2openfoundation,zawalski2024roboticcontrolembodiedchainofthought,liu2023llmpempoweringlargelanguage,xie2023translating}, VLMs provide a distinct advantage by integrating visual data, effectively addressing the perceptual limitations of LLMs in real-world environments.
Notable examples include COME-robot~\cite{zhi2024closed} and VILA~\cite{hu2023look}.
Despite their great potential, VLM-powered methods remain relatively underexplored compared to the extensively studied LLM-powered approaches.
This limited exploration may contribute to the persistent challenges these methods face, particularly in terms of scalability for long-term planning.
To address these issues, ViLaIn~\cite{shirai2024vision} combines VLMs with classical symbolic planning, leveraging Planning Domain Definition Language (PDDL)~\cite{jiang2019task} to refine problem descriptions through error feedback to enhance the accuracy of long-horizon task execution. 
Additionally, DKPROMPT~\cite{zhang2024dkprompt} automates VLM prompting using domain knowledge from PDDL, improving task planning in open-world environments.
Unlike methods that rely solely on traditional planners such as PDDL, our approach, AlignBot, improves scalability by dynamically retrieving contextually relevant examples and integrating them into the prompt, effectively leveraging the in-context learning capabilities of GPT.

\vspace{.5em}
\noindent
\textbf{Contextual Alignment of VLMs in Household Robotics:} Such alignment is crucial, as robots must adapt to complex real-world environments and interact effectively with human users.
Common approaches to achieving this alignment include prompt engineering and fine-tuning.
While prompt engineering is computationally efficient, its success depends on the VLMs' intrinsic capabilities and may not fully achieve the necessary contextual alignment.
Fine-tuning can enhance performance but requires considerable computational resources and extensive annotated data.
A comparative study by Akiyama et al.~\cite{akiyama2024open} evaluates these two kinds of methods in the context of VLM-powered robotic planning.
Most existing research emphasizes leveraging prompt engineering to align VLMs with real-world environments~\cite{liao2024can,mei2024gamevlm,xu2024collage}.
Beyond environment-based alignment, user reminders are another crucial aspect of contextual alignment. 
Few studies have incorporated user reminders into the VLM alignment process.
The most relevant is LLM-Personalized~\cite{han2024llm}.
This approach, however, remains limited, focusing primarily on individual personalization rather than broader contextual alignment. 
In contrast, AlignBot incorporates a wider range of user reminders, enabling more adaptive and robust robotic task planning.

\begin{figure*}[t]
    \centering
    \includegraphics[width=0.82\linewidth]{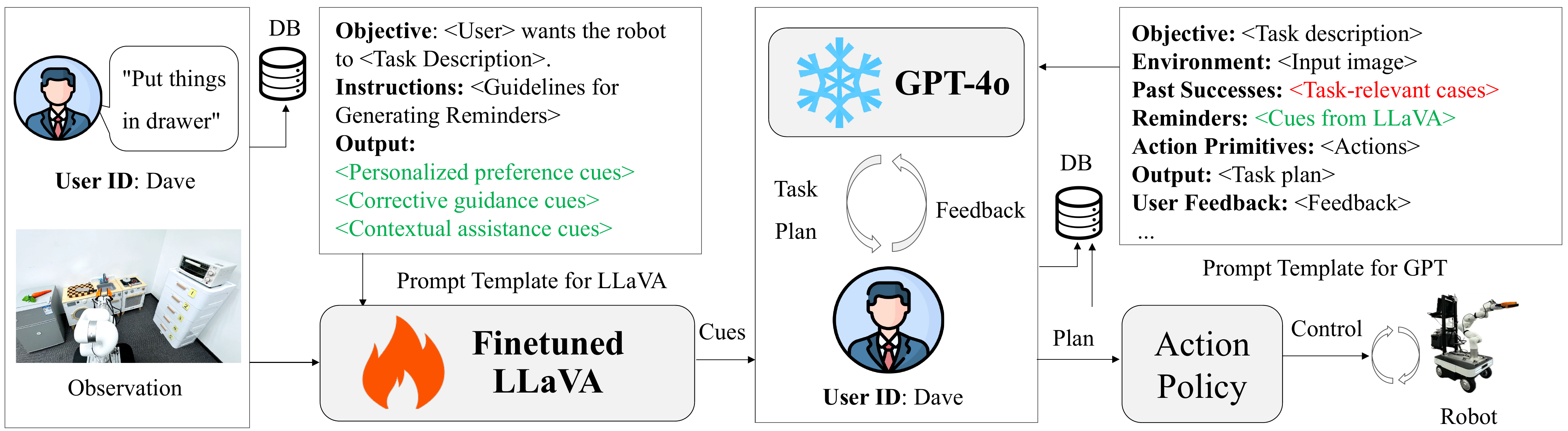}
    \vspace{-0.5em}
    \caption{The fine-tuned \emph{LLaVA model} serves as an adapter for \emph{GPT-4o} during inference, processing user id, task descriptions, and observations to produce \textcolor{mygreen}{cues} that guide GPT-4o's task planning.
    These cues, combined with a dynamically retrieved \textcolor{myred}{task-relevant cases} of past successes, are incorporated into the prompt, optimizing GPT-4o's generation of action plans.
    If the initial output does not meet user expectations, the system enables iterative dialogue, supporting multiple rounds of feedback and refinement until a satisfactory result is achieved.
    \vspace{-1.5em}}
    \label{fig:framework_zx}
\end{figure*}

\section{Problem Formulation}


In this work, we address the problem of aligning robotic customized task planning with customized user reminders.
Traditional task planning for robots typically involves generating action sequences to accomplish a specified task within a given environment, often without considering user reminders. 
In contrast, our problem setting focuses on a realistic scenario where a robot operates in a household with multiple users and is continuously evaluated on its task planning capabilities.
This scenario introduces user id and their reminders directly into the task planning framework.

For each task planning problem $i$, the system receives as \textbf{input} a unique user identifier $u_j$ corresponding to one of the household users, observations $o_i$ of the specific task environment, and a task description $t_i$ provided by the user.
The complete set of input parameters is denoted as $\langle u_j, o_i, t_i \rangle$.
The system \textbf{outputs} a sequence of actions, denoted as $p_i$, aimed at fulfilling the task specified by $t_i$ within the environment described as $o_i$.
The output is considered \emph{successful} when the task description $t_i$ is achieved, and the user expresses satisfaction with the action plan $p_i$, offering no further corrective feedback.
The system also facilitates multi-round dialogue $q_i$, enabling continuous interaction and iterative task-specific user reminders during and after execution.
As the robot continues to execute tasks, successful completions, along with their corresponding multimodal inputs, are stored as historical data, denoted as $D$, where each entry is denoted as $d_i=\langle u_j, o_i, t_i, q_i, p_i\rangle$.
This accumulated dataset forms the foundation for continuous optimization, allowing the system to progressively refine its task planning capabilities.
We \textbf{assume} the presence and accessibility of all objects required for task execution within the environment, ensuring that the robot is able to complete tasks are instructed.
The \textbf{objective} is to maximize the success rate of task completion by effectively aligning the robot's actions with user reminders, thereby improving performance across successive tasks.

\section{The AlignBot Approach}
Fig.~\ref{fig:framework_zx} illustrates the architecture of AlignBot, which comprises two primary components: a fine-tuned LLaVA and GPT-4o. 
The LLaVA model generates instruction-formatted cues to guide GPT-4o in task planning, ensuring alignment with customized user reminders.
Each user-GPT interaction-such as user inputs and GPT responses-is stored in a multimodal database. 
This historical data is used to continually fine-tune LLaVA, enabling the system to improve over time by learning from past interactions and aligning more effectively with user-specific reminders.
AlignBot's effectiveness is built upon two key techniques: (1) the fine-tuning of LLaVA to better align with user reminders; and (2) a case-based learning approach that refines GPT-4o's prompting by integrating relevant historically successful action plans, specifically designed to address action omissions and sequence errors in complex, long-horizon tasks.

\subsection{Fine-Tuning LLaVA with User Reminders}

\vspace{.5em}
\noindent
\textbf{User Reminder:}
The raw dataset, denoted as $D$, is collected through a structured data pipeline, which systematically captures detailed user-GPT interactions.
Each dialogue $q$ in the dataset encapsulates task-specific user reminders on the generated task plans. 
This reminder is typically classified into three categories~\cite{ren2023robotsaskhelpuncertainty,liu2023reflectsummarizingrobotexperiences,ding2023integrating,han2024llm,wu2023tidybot}.
\emph{Personalized preferences} reflect the unique needs of a user, such as a household requiring the robot to provide only sugar-free drinks for a family member with diabetes. 
\emph{Corrective guidance} refers to adjustments provided by users when the robot's plan contains errors, such as omitting the step to turn off the stove after cooking.
\emph{Contextual assistance} is invoked when the robot needs help interpreting its environment, such as correctly identifying hazardous materials (e.g., a bottle of household cleaner) that require special handling.

\begin{figure}[t]
    \centering
    \includegraphics[width=1\linewidth]{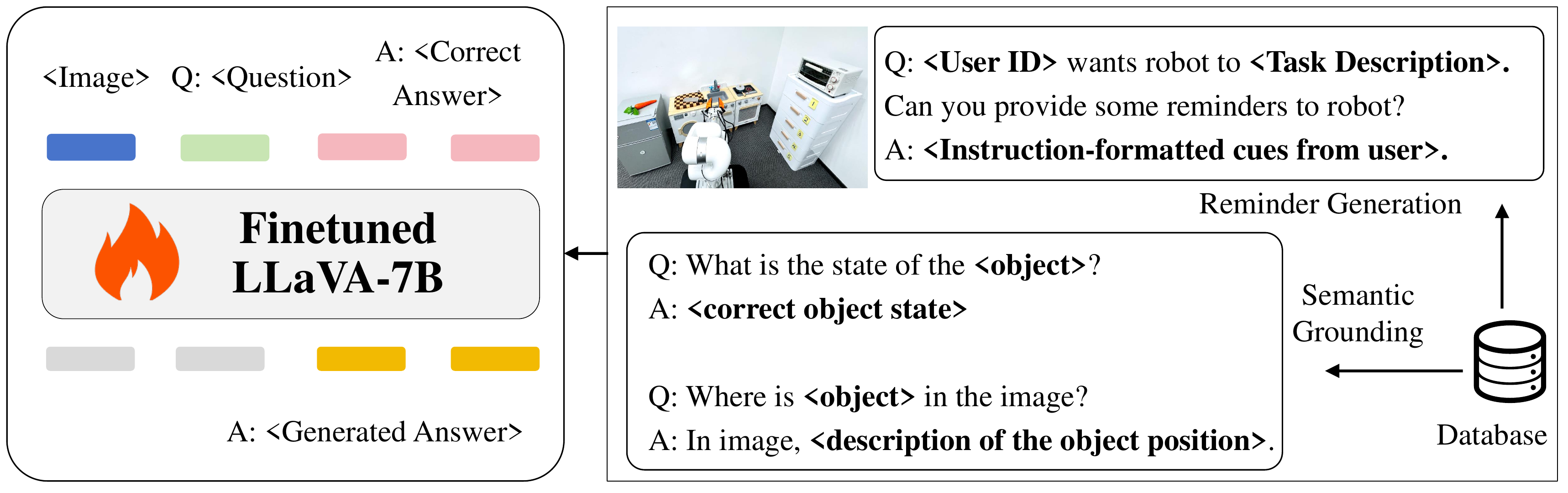}
    \caption{Illustration of fine-tuning LLaVA.
    LLaVA is fine-tuned to optimize both semantic grounding and cue generation.
    The dataset is organized in a Question-Answer (Q\&A) format. 
    During training, LLaVA receives an image, a question, and the corresponding correct answer as input, and the output is the generated response.
    \vspace{-1.5em}}
    \label{fig:finetune}
\end{figure}

\vspace{.5em}
\noindent
\textbf{Datasets for Finetuning:}
From the raw dataset $D$, two specialized training datasets are derived:
one for enhancing LLaVA's semantic grounding and the other for generating cues to be used as prompts for GPT-4o.
\emph{The need for semantic grounding emerges from instances where LLaVA exhibited difficulties in accurately recognizing objects and their states as described in user reminders, leading to misalignment in task execution.}
Both datasets are constructed in a structured Question-Answer (Q\&A) format, as shown in Fig.~\ref{fig:finetune} (right side).
The \emph{semantic grounding dataset}, referred to as $D_s$, addresses two core aspects: task-relevant object recognition and state recognition.
For object recognition, multiple images of each object are sourced from the database, covering a range of positions, angles, and distances within the scene, as well as instances where objects are partially occluded.
Image augmentation techniques, such as cropping, are applied where necessary to ensure comprehensive data coverage.
For state recognition, the dataset includes visual examples depicting objects in different states (e.g., a refrigerator door is open or closed), under varying lighting conditions, angles, and partial occlusions.
By drawing from a diverse set of visual representations, the $D_s$ dataset serves as a comprehensive resource, enabling LLaVA to more effectively map visual inputs to their corresponding semantic interpretations, thereby improving both object identification and state recognition.
Further details are available in Appendix A\footnote{Appendix A: \url{https://yding25.com/AlignBot/Appendix_A}}

The \emph{cue generation dataset}, denoted as $D_c$, consists of structured entries, each comprising an observation captured during task execution, along with the task description, user identifier, and extracted human reminder in the form of the instruction-formatted cue.
These cues, representing user reminders distilled from the dialogues $q_i$, are collected manually.
In cases where multiple reminder instances appear within a single dialogue, these entries are consolidated into one, ensuring that LLaVA generates all necessary cues in a single inference step, thereby accommodating the cumulative reminder typical of multi-round user interactions.
The dataset was generated from three primary scenarios: kitchen, living room, and tabletop settings, with the objects spanning a wide range (refer to  Appendix A for details).
To enhance GPT-4o's comprehension of the cues generated by LLaVA, we formulated specific guidelines from practical experience.
These guidelines prioritize clarity, contextual relevance, and precise execution, as detailed in Appendix B\footnote{Appendix B: \url{https://yding25.com/AlignBot/Appendix_B}}.

\vspace{.5em}
\noindent
\textbf{LLaVA Fine-Tuning:}
For fine-tuning, we employ the LLaVA-7B model\footnote{https://huggingface.co/llava-hf/llava-1.5-7b-hf}, integrating it with GPT-4o by freezing the GPT-4o parameters and exclusively fine-tuning LLaVA.
This approach preserves GPT's task planning capabilities while enabling LLaVA to achieve better performance in both semantic grounding and cue generation tasks.
LLaVA-7B is selected over LLaVA-13B due to its lower data requirements, rendering it more suitable for contexts with limited training data.
Although LLaVA-7B exhibits comparatively weaker reasoning capabilities than LLaVA-13B, the latter demonstrates no significant performance gains when trained on small datasets, thereby making LLaVA-7B a more computationally efficient choice under our constraints. 
To further optimize LLaVA-7B's fine-tuning, we leverage the LoRA method~\cite{hu2022lora}. 
This enables efficient fine-tuning by reducing the number of trainable parameters, allowing the model to swiftly capture critical details even in low-data settings.

We identify two distinct fine-tuning objectives: semantic grounding and cue generation.
Initial attempts at joint training under a single loss function are abandoned due to task interference, as the objectives of semantic grounding and cue generation inherently diverge. 
Consequently, we adopt a phased fine-tuning strategy, initially concentrating on semantic grounding before transitioning to cue generation. 
This sequential approach allows the model to build upon the knowledge gained during the semantic grounding phase, such as object and state recognition, thereby enhancing its capacity to generate more accurate cues aligned with user reminders.
We further observe that extended prompts negatively impacted the performance of the LLaVA-7B model. 
To mitigate this, we design concise prompts, such as:
\emph{\textless User ID\textgreater ~wants a robot to \textless Task Description\textgreater. Can you provide some cues to the robot?}
This approach ensured that the prompts remained clear and focused on the task at hand, minimizing the risk of performance degradation caused by overly complex inputs.
During inference, we incorporate zero-shot chain-of-thought reasoning by adding the prompt ``Let's think step by step", which guides the model to generate more coherent and detailed reasoning in its responses. 
This helps maintain the model's reasoning abilities without relying on lengthy prompts that could reduce clarity.

The fine-tuning process is optimized with the Cross-Entropy Loss function~\cite{hu2022lora,gao2024learning}.
The number of epochs is carefully set to 40 for our fine-tuning process, representing a substantial increase compared to the majority of existing studies, which typically employ fewer than 10 epochs.
This decision is motivated by the specific context of our work: a single household environment with task repetition and user-specific details. 
Although increasing the number of epochs can reduce generalization, this trade-off is acceptable in our case, as the model needed to retain detailed information for long-term adaptation to a particular user.
We also explore the cold-start problem, particularly when working with very small datasets (e.g., dozens of samples). 
In such cases, we increase the number of epochs to 60 to help the model memorize the initial reminder. 
As the dataset size expands, we gradually reduce the epochs back to 40 to ensure the model can adapt to new tasks without losing the ability to recall past interactions. 
We experimented with learning rates ranging from $1 \times 10^{-5}$ to $5 \times 10^{-5}$ and found no significant differences in performance across this range.
Further details on hyperparameters are provided in Appendix B.

\subsection{Case-Based Learning for Enhanced GPT Prompting}
Long-horizon task planning in household robotics frequently encounters issues such as action omissions and sequence errors, undermining execution reliability. 
For example, a robot may attempt to place an object without first executing the necessary action of grasping it.
To address these challenges, we propose a \emph{case-based learning} approach that enhances GPT-4o’s planning capabilities by incorporating relevant historical cases into the prompting process.
This method involves dynamically retrieving the top-$k$ most relevant cases-specifically, past action plans that were both successful and approved by users-from historical dialogue data.
Here, we refer to these as cases.
These cases provide GPT-4o with contextual guidance from past successes.
By using similar, correct past action sequences, we aim to improve plan accuracy and robustness.
The rationale is that past successful plans contain effective solutions to prevent common errors.
However, retrieving relevant cases in a multimodal context poses challenges due to the complexity of assessing similarity across different modalities~\cite{li2024survey}.
To overcome this, we first select successful plans that match the same task description. 
We then utilize GPT-4o's feedback to evaluate and rank these plans based on their relevance and effectiveness. 
By integrating the most appropriate cases, GPT-4o generates more accurate and comprehensive action sequences, reducing omissions and improving execution.

Consider a subset $D_r$ of raw dataset $D$, defined as $D_r = \{\langle p_1,t_1 \rangle, \langle p_2,t_2 \rangle, \dots\}$, where each action plan $p_i$ is linked to a historical corresponding task description $t_i$. 
Each plan $p_i$ is assigned an initial priority score $f_i = f_0$.
When a new task $t_{\text{new}}$ is introduced, the system selects a subset $D_r' \subset D_r$ of action plans relevant to $t_{\text{new}}$, determined by the similarity function $sim(t_i, t_{\text{new}})$.
Specifically, $D_r' = \{ p_i \mid sim(t_i, t_{\text{new}}) > \tau \}$,
where $\tau$ is a predefined similarity threshold. 
The similarity is computed using TF-IDF~\cite{sparck1972statistical} and cosine similarity.
The action plans in $D_r'$ are then sorted based on their priority scores $f_i$, resulting in an ordered set.
To normalize these priority scores and prevent over-reliance on a limited set of action plans, a Softmax function is applied:
$f_{i_j} = \exp(\hat{f}_{i_j}) / \sum_{k=1}^{|S|} \exp(\hat{f}_{i_k})$.
To dynamically adjust priorities based on effectiveness, we introduce a \emph{gradient-based update mechanism}.
Each plan $p_i$ has a gradient $\Delta f$, indicating its priority change rate.
The gradient is updated depending on the utility of the action plan in the current task:
$\Delta f_{\text{positive}} = \Delta f_{\text{positive}}^0 \cdot \exp(-\alpha \cdot n_{i_j})$,
$\Delta f_{\text{negative}} = \Delta f_{\text{negative}}^0 \cdot \exp(-\beta \cdot n_{i_j})$, 
where $\alpha$ and $\beta$ are decay factors controlling how quickly priorities change with repeated use, and $n_{i_j}$ denotes the usage frequency of plan $p_i$.
The initial gradient $\Delta f^0$ is a fixed value controlling the amplitude of updates, preventing unbounded growth or decay of priority scores over time. 
After updating the gradient, the priority score $f_i$ is adjusted accordingly:
$ f_i = \min(\max(f_i + \Delta f, 0.1), 1.0) $.
The selection of action plans is governed by an $\epsilon$-Greedy strategy to balance exploration and exploitation.
Specifically, with probability $\epsilon$, a random action plan $F_{\text{random}}$ is chosen to encourage exploration, while with probability $1 - \epsilon$, the highest-priority plan $F_{\text{max}}$ is selected to exploit known effective strategies.
The parameter $\epsilon$ controls this balance, ensuring adaptability without over-reliance on a limited set of strategies. 
By iteratively applying this method, the system refines its task planning, continuously adapting to user preferences and specific task contexts. 
Further details can be found in Appendix B.

\subsection{Action Policy for Task Execution}
After GPT outputs the task plan and receives user approval, the robot proceeds to execute each action in sequence, such as opening a drawer. 
To execute complex actions such as opening drawers, we employ the ACT algorithm~\cite{zhao2023learningfinegrainedbimanualmanipulation}, while simpler tasks like pick-and-place operations are handled using the AnyGrasp method~\cite{fang2023anygrasprobustefficientgrasp} to achieve more efficient execution. 
For more technical details, see Appendix B.

\section{Experiments}
We assess the performance of AlignBot from two key perspectives: the quality of task plans generated by GPT-4o and the quality of cues produced by the fine-tuned LLaVA.

\vspace{.5em}
\noindent
\textbf{Baselines:} As no existing methods closely match our approach for direct comparison, we designed three baselines to systematically evaluate AlignBot's performance. 
\begin{itemize}
    \item \emph{\textbf{Vanilla GPT-4o}}:  
    In this baseline, GPT-4o is used solely for task planning, relying only on the given task description and observation. 
    Unlike AlignBot, it does not utilize cues generated by the fine-tuned LLaVA or incorporate dynamic retrieval of task-relevant cases.
    This setup serves as a fundamental comparison point to evaluate the standalone performance of GPT-4o without additional enhancements.
    \item \emph{\textbf{GPT-4o + Raw Reminder}:}
    This baseline investigates the impact of integrating GPT-4o with randomly retrieved, unprocessed user-robot interaction data.
    These interactions correspond to the cues and task-relevant cases used in AlignBot but are not structured or instruction-formatted.
    These raw interactions, which may include irrelevant or conflicting information, are directly incorporated into GPT-4o's prompts. 
    This baseline evaluates the significance of LLaVA-powered cue generation and relevance filtering in enhancing GPT-4o's task planning capabilities, as implemented in AlignBot.
    \item \emph{\textbf{GPT-4o + Fine-tuned LLaMA}:}
    This baseline evaluates the effectiveness of using a \textbf{single-modal} language model to generate cues for GPT-4o.
    Specifically, LLaMA2-7b~\cite{touvron2023llama2openfoundation} is fine-tuned solely on text data, generating text-based cues for the goal. 
    GPT-4o then uses these cues to generate a plan. 
    By comparing performance when visual data is excluded, this setup highlights the importance of multimodal models.
\end{itemize}

\begin{figure*}[t]
    \centering
    \includegraphics[width=0.8\linewidth]{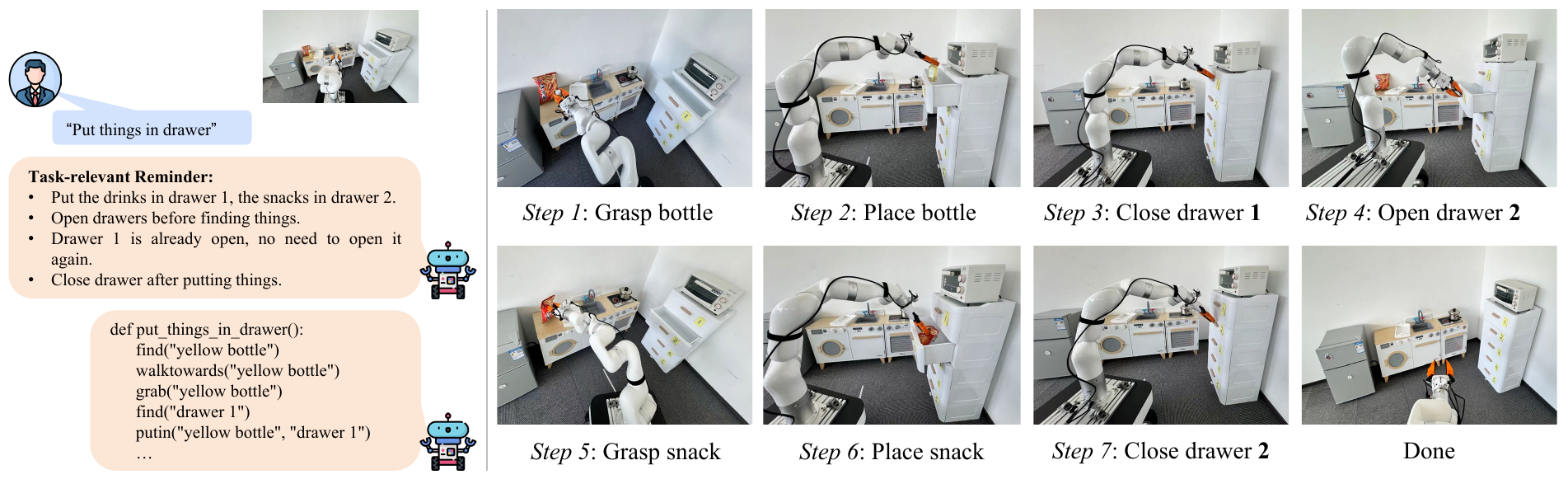}
    \vspace{-0.8em}
    \caption{\textbf{Real Robot Demonstration.} We implement AlignBot on a real robotic system, comprising an AgileX-based mobile platform and a UFactory XArm robotic arm.
    The system employs the ACT algorithm alongside the AnyGrasp method for manipulation. 
    In this setup, the robot is tasked with placing items from a countertop into a drawer, ensuring that the task plan generated is consistently aligned with user reminders.
    \vspace{-1.5em}
    }
    \label{fig:framework}
\end{figure*}

\vspace{.5em}
\noindent
\textbf{Experimental Setup:}
We develop a benchmark dataset in a controlled laboratory environment, constructing three distinct scenes: a kitchen, a living room, and a tabletop setting. 
These settings encompass a total of 20 distinct everyday tasks, such as organizing fruits into a bowl. 
Half of the tasks replicate those from prior work~\cite{liu2023reflectsummarizingrobotexperiences}, while the other half are newly designed for this study. 
The tasks involve six categories of objects, including kitchen utensils and fruits, comprising over 80 unique items. 
Task complexity varies, ranging from simple to long-term, multi-step tasks. 
Each task scene consists of 2–6 operative objects and 1–3 non-target objects, with variations in object position, state, and viewing angle. 
Approximately 50–150 images are captured per task, resulting in a total of over 1,500 task scenes for training LLaVA, where three volunteers are involved in interacting with GPT-4o.
Separately, for the evaluation of AlignBot, we generate 600 task executions from 20 tasks across 5 scenarios per task, varying in object positions, states, and distractor items. 
The same three volunteers perform each task twice, providing a total of 600 evaluations. 
These volunteers assess whether the generated plans are reasonable and whether task goals are successfully achieved. 
Further details are provided in Appendix C\footnote{Appendix C: \url{https://yding25.com/AlignBot/Appendix_C}}.

\vspace{.5em}
\noindent
\textbf{Rating Criteria:} 
We assess the methods' effectiveness by comparing their planning \textbf{success rates}, where the methods are anonymized to volunteers for fair evaluation. 
To evaluate the quality of the generated cues, we select two scenes for each task, and the volunteers rate the cues without knowing which method generates them, ensuring objectivity.
The \textbf{user rating} scale ranges from 0 to 3, with higher scores indicating better cue quality. 
The rating guidelines for human raters in the experiments are available in Appendix C.

\vspace{.5em}
\noindent
\textbf{AlignBot vs. Baselines:} TABLE~\ref{tab:main} shows the comparison result between AlignBot and the three baselines.
AlignBot outperforms all baselines, demonstrating superior performance in aligning task planning with customized user reminders.
Compared to \emph{Vanilla GPT-4o}, \emph{AlignBot} addresses significant limitations in semantic grounding, such as recognizing the object states (e.g., whether a drawer is open or closed). 
\emph{Vanilla GPT-4o} tends to make random decisions in tasks due to missing user-specific information, while AlignBot, through its database of user reminders and fine-tuned cues, delivers much more accurate and effective task plans. 
When compared to \emph{GPT-4o + Raw Reminder}, AlignBot again proves more effective, as \emph{GPT-4o + Raw Reminder} often struggles to filter and apply the correct historical data for the current scenario, resulting in planning failures. 
The context-specific cues generated by AlignBot's adapter are essential for accurate task planning, particularly in complex environments.
Finally, while \emph{GPT-4o + Fine-tuned LLaMA} can remember user preferences and correct some common planning errors, its lack of multimodal capabilities limits its understanding of the scene and the arrangement of objects, leading to less reliable cues. 
AlignBot's ability to integrate multimodal inputs ensures better scene awareness and task alignment, demonstrating the necessity of multimodal adapters in improving task planning.

\begin{table}[thp]
\centering
\caption{Overall performance of AlignBot as compared to three baselines based on the success rates.}\label{tab:main}
\begin{tabular}{l|c}
\toprule
    \textbf{Task Planning Methods} & \textbf{Average Success Rate}\\
    \midrule
    \emph{Vanilla GPT-4o} & 21.67\%  \\
    \midrule
    \emph{GPT-4o + Raw Reminder} & 33.15\%\\
    \midrule
    \emph{GPT-4o + Fine-tuned LLaMA} & 40.19\%\\
    \midrule
    \emph{AlignBot (ours)} & \textbf{86.85}\%\\
    \bottomrule
\end{tabular}
\vspace{-1em}
\end{table}

\begin{table}[thp]
\centering
\caption{Comparison between three cue generators based on user rating (with a maximum score of 3).}
\vspace{-0.5em}
\begin{tabular}{l|c}
\toprule
    \textbf{Cue Generator} & \textbf{Average User Rating}\\
    \midrule
    \emph{Fine-tuned LLaMA} & 0.54\\
    \midrule
    \emph{LLaVA without Fine-Tuning} & 1.18\\
    \midrule
    \emph{Fine-tuned LLaVA (ours)} & \textbf{2.75}\\
    \bottomrule
\end{tabular}
\vspace{-1.5em}
\end{table}

\vspace{.5em}
\noindent
\textbf{Quality of Cues:}
Our AlignBot demonstrates superior performance in cue generation compared to baselines.
It receives high ratings from volunteers for its ability to combine task descriptions with image context, which allows it to deliver highly effective, scene-based cues. 
It also excels at remembering personalized user preferences, making its responses more accurate and tailored to individual needs, which significantly enhances its utility in real-world applications. 
When compared to \emph{Fine-tuned LLaMA}, AlignBot's multimodal capabilities are a clear advantage. 
While \emph{Fine-tuned LLaMA} can remember user preferences, its single-modal nature limits its effectiveness, especially in tasks requiring visual input.
Without image guidance, \emph{Fine-tuned LLaMA} randomly selects from past reminders and fails to provide accurate or context-specific cues, underscoring the importance of multimodal inputs in personalized tasks. 
In contrast, the \emph{LLaVA without Fine-tuning} further highlights AlignBot's strengths, as it struggles with image-based reasoning often providing vague or generic descriptions of task-relevant items.
\emph{LLaVA without Fine-tuning} lacks the ability to interpret detailed visual information or infer operational errors, making it far less effective in guiding robots through real-world tasks.
Additionally, without fine-tuning to incorporate user-specific data, its responses remain generic and less relevant, reinforcing the importance of AlignBot's tailored and fine-tuned approach to cue generation.

\section{Conclusion}
To summarize, AlignBot demonstrates significant improvements in aligning robotic task planning with diverse, multimodal user reminders. 
By fine-tuning LLaVA-7B as an adapter for GPT-4o, the framework effectively handles the challenges posed by the nature of user reminders in household environments. 
The combination of instruction-formatted cues and case-based learning enables AlignBot to generate more accurate task plans. 
Empirical results show that AlignBot outperforms baselines.

\section*{Acknowledgment}
This work is supported by the Shanghai AI Laboratory, the National Key R\&D Program of China (2022ZD0160102), the National Natural Science Foundation of China (62376222), and the Young Elite Scientists Sponsorship Program by CAST (2023QNRC001).

\bibliographystyle{IEEEtran}
\bibliography{IEEEabrv,references}

\begin{thebibliography}{10}
\providecommand{\url}[1]{#1}
\csname url@rmstyle\endcsname
\providecommand{\newblock}{\relax}
\providecommand{\bibinfo}[2]{#2}
\providecommand\BIBentrySTDinterwordspacing{\spaceskip=0pt\relax}
\providecommand\BIBentryALTinterwordstretchfactor{4}
\providecommand\BIBentryALTinterwordspacing{\spaceskip=\fontdimen2\font plus
\BIBentryALTinterwordstretchfactor\fontdimen3\font minus \fontdimen4\font\relax}
\providecommand\BIBforeignlanguage[2]{{%
\expandafter\ifx\csname l@#1\endcsname\relax
\typeout{** WARNING: IEEEtran.bst: No hyphenation pattern has been}%
\typeout{** loaded for the language `#1'. Using the pattern for}%
\typeout{** the default language instead.}%
\else
\language=\csname l@#1\endcsname
\fi
#2}}

\bibitem{ding2023task}
Y.~Ding, X.~Zhang, C.~Paxton, and S.~Zhang, ``Task and motion planning with large language models for object rearrangement,'' in \emph{2023 IEEE/RSJ International Conference on Intelligent Robots and Systems (IROS)}.\hskip 1em plus 0.5em minus 0.4em\relax IEEE, 2023, pp. 2086--2092.

\bibitem{ding2023integrating}
Y.~Ding, X.~Zhang, S.~Amiri, N.~Cao, H.~Yang, A.~Kaminski, C.~Esselink, and S.~Zhang, ``Integrating action knowledge and llms for task planning and situation handling in open worlds,'' \emph{Autonomous Robots}, vol.~47, no.~8, pp. 981--997, 2023.

\bibitem{wu2023tidybot}
J.~Wu, R.~Antonova, A.~Kan, M.~Lepert, A.~Zeng, S.~Song, J.~Bohg, S.~Rusinkiewicz, and T.~Funkhouser, ``Tidybot: Personalized robot assistance with large language models,'' \emph{Autonomous Robots}, vol.~47, no.~8, pp. 1087--1102, 2023.

\bibitem{zhang2023hierarchical}
M.~Zhang, G.~Tian, Y.~Cui, Y.~Zhang, and Z.~Xia, ``Hierarchical semantic knowledge-based object search method for household robots,'' \emph{IEEE Transactions on Emerging Topics in Computational Intelligence}, 2023.

\bibitem{driess2023palmeembodiedmultimodallanguage}
\BIBentryALTinterwordspacing
D.~Driess, F.~Xia, M.~S.~M. Sajjadi, C.~Lynch, A.~Chowdhery, B.~Ichter, A.~Wahid, J.~Tompson, Q.~Vuong, T.~Yu, W.~Huang, Y.~Chebotar, P.~Sermanet, D.~Duckworth, S.~Levine, V.~Vanhoucke, K.~Hausman, M.~Toussaint, K.~Greff, A.~Zeng, I.~Mordatch, and P.~Florence, ``Palm-e: An embodied multimodal language model,'' 2023. [Online]. Available: \url{https://arxiv.org/abs/2303.03378}
\BIBentrySTDinterwordspacing

\bibitem{vemprala2023chatgptroboticsdesignprinciples}
\BIBentryALTinterwordspacing
S.~Vemprala, R.~Bonatti, A.~Bucker, and A.~Kapoor, ``Chatgpt for robotics: Design principles and model abilities,'' 2023. [Online]. Available: \url{https://arxiv.org/abs/2306.17582}
\BIBentrySTDinterwordspacing

\bibitem{huang2024rekepspatiotemporalreasoningrelational}
\BIBentryALTinterwordspacing
W.~Huang, C.~Wang, Y.~Li, R.~Zhang, and L.~Fei-Fei, ``Rekep: Spatio-temporal reasoning of relational keypoint constraints for robotic manipulation,'' 2024. [Online]. Available: \url{https://arxiv.org/abs/2409.01652}
\BIBentrySTDinterwordspacing

\bibitem{Lin_2023}
\BIBentryALTinterwordspacing
K.~Lin, C.~Agia, T.~Migimatsu, M.~Pavone, and J.~Bohg, ``Text2motion: from natural language instructions to feasible plans,'' \emph{Autonomous Robots}, vol.~47, no.~8, p. 1345–1365, Nov. 2023. [Online]. Available: \url{http://dx.doi.org/10.1007/s10514-023-10131-7}
\BIBentrySTDinterwordspacing

\bibitem{Wake_2023}
\BIBentryALTinterwordspacing
N.~Wake, A.~Kanehira, K.~Sasabuchi, J.~Takamatsu, and K.~Ikeuchi, ``Chatgpt empowered long-step robot control in various environments: A case application,'' \emph{IEEE Access}, vol.~11, p. 95060–95078, 2023. [Online]. Available: \url{http://dx.doi.org/10.1109/ACCESS.2023.3310935}
\BIBentrySTDinterwordspacing

\bibitem{ahn2024autortembodiedfoundationmodels}
\BIBentryALTinterwordspacing
M.~Ahn, D.~Dwibedi, C.~Finn, M.~G. Arenas, K.~Gopalakrishnan, K.~Hausman, B.~Ichter, A.~Irpan, N.~Joshi, R.~Julian, S.~Kirmani, I.~Leal, E.~Lee, S.~Levine, Y.~Lu, I.~Leal, S.~Maddineni, K.~Rao, D.~Sadigh, P.~Sanketi, P.~Sermanet, Q.~Vuong, S.~Welker, F.~Xia, T.~Xiao, P.~Xu, S.~Xu, and Z.~Xu, ``Autort: Embodied foundation models for large scale orchestration of robotic agents,'' 2024. [Online]. Available: \url{https://arxiv.org/abs/2401.12963}
\BIBentrySTDinterwordspacing

\bibitem{driess2023palm}
D.~Driess, F.~Xia, M.~S. Sajjadi, C.~Lynch, A.~Chowdhery, B.~Ichter, A.~Wahid, J.~Tompson, Q.~Vuong, T.~Yu, \emph{et~al.}, ``Palm-e: An embodied multimodal language model,'' \emph{arXiv preprint arXiv:2303.03378}, 2023.

\bibitem{zhu2023minigpt}
D.~Zhu, J.~Chen, X.~Shen, X.~Li, and M.~Elhoseiny, ``Minigpt-4: Enhancing vision-language understanding with advanced large language models,'' \emph{arXiv preprint arXiv:2304.10592}, 2023.

\bibitem{zhou2024proreasonmultimodalproactivereasoning}
\BIBentryALTinterwordspacing
J.~Zhou, S.~Wang, J.~Dong, L.~Li, J.~Gao, L.~Kong, and C.~Wu, ``Proreason: Multi-modal proactive reasoning with decoupled eyesight and wisdom,'' 2024. [Online]. Available: \url{https://arxiv.org/abs/2410.14138}
\BIBentrySTDinterwordspacing

\bibitem{openai2024gpt4o}
OpenAI, ``Gpt-4o: A multimodal task planning model,'' \url{https://openai.com/index/hello-gpt-4o/}, 2024.

\bibitem{ren2023robotsaskhelpuncertainty}
\BIBentryALTinterwordspacing
A.~Z. Ren, A.~Dixit, A.~Bodrova, S.~Singh, S.~Tu, N.~Brown, P.~Xu, L.~Takayama, F.~Xia, J.~Varley, Z.~Xu, D.~Sadigh, A.~Zeng, and A.~Majumdar, ``Robots that ask for help: Uncertainty alignment for large language model planners,'' 2023. [Online]. Available: \url{https://arxiv.org/abs/2307.01928}
\BIBentrySTDinterwordspacing

\bibitem{liu2023reflectsummarizingrobotexperiences}
\BIBentryALTinterwordspacing
Z.~Liu, A.~Bahety, and S.~Song, ``Reflect: Summarizing robot experiences for failure explanation and correction,'' 2023. [Online]. Available: \url{https://arxiv.org/abs/2306.15724}
\BIBentrySTDinterwordspacing

\bibitem{han2024llm}
D.~Han, T.~McInroe, A.~Jelley, S.~V. Albrecht, P.~Bell, and A.~Storkey, ``Llm-personalize: Aligning llm planners with human preferences via reinforced self-training for housekeeping robots,'' \emph{arXiv preprint arXiv:2404.14285}, 2024.

\bibitem{shi2024yellrobotimprovingonthefly}
\BIBentryALTinterwordspacing
L.~X. Shi, Z.~Hu, T.~Z. Zhao, A.~Sharma, K.~Pertsch, J.~Luo, S.~Levine, and C.~Finn, ``Yell at your robot: Improving on-the-fly from language corrections,'' 2024. [Online]. Available: \url{https://arxiv.org/abs/2403.12910}
\BIBentrySTDinterwordspacing

\bibitem{zhi2024closed}
P.~Zhi, Z.~Zhang, M.~Han, Z.~Zhang, Z.~Li, Z.~Jiao, B.~Jia, and S.~Huang, ``Closed-loop open-vocabulary mobile manipulation with gpt-4v,'' \emph{arXiv preprint arXiv:2404.10220}, 2024.

\bibitem{hu2023look}
Y.~Hu, F.~Lin, T.~Zhang, L.~Yi, and Y.~Gao, ``Look before you leap: Unveiling the power of gpt-4v in robotic vision-language planning,'' \emph{arXiv preprint arXiv:2311.17842}, 2023.

\bibitem{liao2024can}
Y.-H. Liao, R.~Mahmood, S.~Fidler, and D.~Acuna, ``Can feedback enhance semantic grounding in large vision-language models?'' \emph{arXiv preprint arXiv:2404.06510}, 2024.

\bibitem{mei2024gamevlm}
A.~Mei, J.~Wang, G.-N. Zhu, and Z.~Gan, ``Gamevlm: A decision-making framework for robotic task planning based on visual language models and zero-sum games,'' \emph{arXiv preprint arXiv:2405.13751}, 2024.

\bibitem{rafailov2024directpreferenceoptimizationlanguage}
\BIBentryALTinterwordspacing
R.~Rafailov, A.~Sharma, E.~Mitchell, S.~Ermon, C.~D. Manning, and C.~Finn, ``Direct preference optimization: Your language model is secretly a reward model,'' 2024. [Online]. Available: \url{https://arxiv.org/abs/2305.18290}
\BIBentrySTDinterwordspacing

\bibitem{Kim_2024}
\BIBentryALTinterwordspacing
Y.~Kim, D.~Kim, J.~Choi, J.~Park, N.~Oh, and D.~Park, ``A survey on integration of large language models with intelligent robots,'' \emph{Intelligent Service Robotics}, Aug. 2024. [Online]. Available: \url{http://dx.doi.org/10.1007/s11370-024-00550-5}
\BIBentrySTDinterwordspacing

\bibitem{liu2024humanawarenessrobottask}
\BIBentryALTinterwordspacing
Y.~Liu, L.~Palmieri, S.~Koch, I.~Georgievski, and M.~Aiello, ``Towards human awareness in robot task planning with large language models,'' 2024. [Online]. Available: \url{https://arxiv.org/abs/2404.11267}
\BIBentrySTDinterwordspacing

\bibitem{liu2023improvedllava}
H.~Liu, C.~Li, Y.~Li, and Y.~J. Lee, ``Improved baselines with visual instruction tuning,'' 2023.

\bibitem{hu2022lora}
\BIBentryALTinterwordspacing
E.~J. Hu, Y.~Shen, P.~Wallis, Z.~Allen-Zhu, Y.~Li, S.~Wang, L.~Wang, and W.~Chen, ``Lo{RA}: Low-rank adaptation of large language models,'' in \emph{International Conference on Learning Representations}, 2022. [Online]. Available: \url{https://openreview.net/forum?id=nZeVKeeFYf9}
\BIBentrySTDinterwordspacing

\bibitem{zhou2024empiricalstudyparameterefficientfinetuning}
\BIBentryALTinterwordspacing
X.~Zhou, J.~He, Y.~Ke, G.~Zhu, V.~Gutiérrez-Basulto, and J.~Z. Pan, ``An empirical study on parameter-efficient fine-tuning for multimodal large language models,'' 2024. [Online]. Available: \url{https://arxiv.org/abs/2406.05130}
\BIBentrySTDinterwordspacing

\bibitem{wang2025mosunleashingparameterefficiency}
\BIBentryALTinterwordspacing
S.~Wang, L.~Chen, P.~Chen, J.~Dong, B.~Xue, J.~Jiang, L.~Kong, and C.~Wu, ``Mos: Unleashing parameter efficiency of low-rank adaptation with mixture of shards,'' 2025. [Online]. Available: \url{https://arxiv.org/abs/2410.00938}
\BIBentrySTDinterwordspacing

\bibitem{wang2024prolorapartialrotationempowers}
\BIBentryALTinterwordspacing
S.~Wang, B.~Xue, J.~Ye, J.~Jiang, L.~Chen, L.~Kong, and C.~Wu, ``Prolora: Partial rotation empowers more parameter-efficient lora,'' 2024. [Online]. Available: \url{https://arxiv.org/abs/2402.16902}
\BIBentrySTDinterwordspacing

\bibitem{ratnayake2023prompting}
H.~Ratnayake and C.~Wang, ``A prompting framework to enhance language model output,'' in \emph{Australasian Joint Conference on Artificial Intelligence}.\hskip 1em plus 0.5em minus 0.4em\relax Springer, 2023, pp. 66--81.

\bibitem{shirai2024vision}
K.~Shirai, C.~C. Beltran-Hernandez, M.~Hamaya, A.~Hashimoto, S.~Tanaka, K.~Kawaharazuka, K.~Tanaka, Y.~Ushiku, and S.~Mori, ``Vision-language interpreter for robot task planning,'' in \emph{2024 IEEE International Conference on Robotics and Automation (ICRA)}.\hskip 1em plus 0.5em minus 0.4em\relax IEEE, 2024, pp. 2051--2058.

\bibitem{zhang2024dkprompt}
X.~Zhang, Z.~Altaweel, Y.~Hayamizu, Y.~Ding, S.~Amiri, H.~Yang, A.~Kaminski, C.~Esselink, and S.~Zhang, ``Dkprompt: Domain knowledge prompting vision-language models for open-world planning,'' \emph{arXiv preprint arXiv:2406.17659}, 2024.

\bibitem{ren2024exploreconfidentefficientexploration}
\BIBentryALTinterwordspacing
A.~Z. Ren, J.~Clark, A.~Dixit, M.~Itkina, A.~Majumdar, and D.~Sadigh, ``Explore until confident: Efficient exploration for embodied question answering,'' 2024. [Online]. Available: \url{https://arxiv.org/abs/2403.15941}
\BIBentrySTDinterwordspacing

\bibitem{singh2022progpromptgeneratingsituatedrobot}
\BIBentryALTinterwordspacing
I.~Singh, V.~Blukis, A.~Mousavian, A.~Goyal, D.~Xu, J.~Tremblay, D.~Fox, J.~Thomason, and A.~Garg, ``Progprompt: Generating situated robot task plans using large language models,'' 2022. [Online]. Available: \url{https://arxiv.org/abs/2209.11302}
\BIBentrySTDinterwordspacing

\bibitem{joublin2023copalcorrectiveplanningrobot}
\BIBentryALTinterwordspacing
F.~Joublin, A.~Ceravola, P.~Smirnov, F.~Ocker, J.~Deigmoeller, A.~Belardinelli, C.~Wang, S.~Hasler, D.~Tanneberg, and M.~Gienger, ``Copal: Corrective planning of robot actions with large language models,'' 2023. [Online]. Available: \url{https://arxiv.org/abs/2310.07263}
\BIBentrySTDinterwordspacing

\bibitem{touvron2023llama2openfoundation}
\BIBentryALTinterwordspacing
H.~Touvron, L.~Martin, K.~Stone, P.~Albert, A.~Almahairi, Y.~Babaei, N.~Bashlykov, S.~Batra, P.~Bhargava, S.~Bhosale, D.~Bikel, L.~Blecher, C.~C. Ferrer, M.~Chen, G.~Cucurull, D.~Esiobu, J.~Fernandes, J.~Fu, W.~Fu, B.~Fuller, C.~Gao, V.~Goswami, N.~Goyal, A.~Hartshorn, S.~Hosseini, R.~Hou, H.~Inan, M.~Kardas, V.~Kerkez, M.~Khabsa, I.~Kloumann, A.~Korenev, P.~S. Koura, M.-A. Lachaux, T.~Lavril, J.~Lee, D.~Liskovich, Y.~Lu, Y.~Mao, X.~Martinet, T.~Mihaylov, P.~Mishra, I.~Molybog, Y.~Nie, A.~Poulton, J.~Reizenstein, R.~Rungta, K.~Saladi, A.~Schelten, R.~Silva, E.~M. Smith, R.~Subramanian, X.~E. Tan, B.~Tang, R.~Taylor, A.~Williams, J.~X. Kuan, P.~Xu, Z.~Yan, I.~Zarov, Y.~Zhang, A.~Fan, M.~Kambadur, S.~Narang, A.~Rodriguez, R.~Stojnic, S.~Edunov, and T.~Scialom, ``Llama 2: Open foundation and fine-tuned chat models,'' 2023. [Online]. Available: \url{https://arxiv.org/abs/2307.09288}
\BIBentrySTDinterwordspacing

\bibitem{zawalski2024roboticcontrolembodiedchainofthought}
\BIBentryALTinterwordspacing
M.~Zawalski, W.~Chen, K.~Pertsch, O.~Mees, C.~Finn, and S.~Levine, ``Robotic control via embodied chain-of-thought reasoning,'' 2024. [Online]. Available: \url{https://arxiv.org/abs/2407.08693}
\BIBentrySTDinterwordspacing

\bibitem{liu2023llmpempoweringlargelanguage}
\BIBentryALTinterwordspacing
B.~Liu, Y.~Jiang, X.~Zhang, Q.~Liu, S.~Zhang, J.~Biswas, and P.~Stone, ``Llm+p: Empowering large language models with optimal planning proficiency,'' 2023. [Online]. Available: \url{https://arxiv.org/abs/2304.11477}
\BIBentrySTDinterwordspacing

\bibitem{xie2023translating}
Y.~Xie, C.~Yu, T.~Zhu, J.~Bai, Z.~Gong, and H.~Soh, ``Translating natural language to planning goals with large-language models,'' \emph{arXiv preprint arXiv:2302.05128}, 2023.

\bibitem{jiang2019task}
Y.-q. Jiang, S.-q. Zhang, P.~Khandelwal, and P.~Stone, ``Task planning in robotics: an empirical comparison of pddl-and asp-based systems,'' \emph{Frontiers of Information Technology \& Electronic Engineering}, vol.~20, pp. 363--373, 2019.

\bibitem{akiyama2024open}
\BIBentryALTinterwordspacing
S.~Akiyama, R.~F.~J. Dossa, K.~Arulkumaran, S.~Sujit, and E.~Johns, ``Open-loop {VLM} robot planning: An investigation of fine-tuning and prompt engineering strategies,'' in \emph{First Workshop on Vision-Language Models for Navigation and Manipulation at ICRA 2024}, 2024. [Online]. Available: \url{https://openreview.net/forum?id=JXngwwPMR5}
\BIBentrySTDinterwordspacing

\bibitem{xu2024collage}
S.~Xu, Y.~Wang, D.~Liu, and C.~Xu, ``Collage prompting: Budget-friendly visual recognition with gpt-4v,'' \emph{arXiv preprint arXiv:2403.11468}, 2024.

\bibitem{gao2024learning}
X.~Gao, P.~Zhang, D.~Qu, D.~Wang, Z.~Wang, Y.~Ding, B.~Zhao, and X.~Li, ``Learning 2d invariant affordance knowledge for 3d affordance grounding,'' \emph{arXiv preprint arXiv:2408.13024}, 2024.

\bibitem{li2024survey}
S.~Li, F.~Huang, and L.~Zhang, ``A survey of multimodal composite editing and retrieval,'' \emph{arXiv preprint arXiv:2409.05405}, 2024.

\bibitem{sparck1972statistical}
K.~Sparck~Jones, ``A statistical interpretation of term specificity and its application in retrieval,'' \emph{Journal of documentation}, vol.~28, no.~1, pp. 11--21, 1972.

\bibitem{zhao2023learningfinegrainedbimanualmanipulation}
\BIBentryALTinterwordspacing
T.~Z. Zhao, V.~Kumar, S.~Levine, and C.~Finn, ``Learning fine-grained bimanual manipulation with low-cost hardware,'' 2023. [Online]. Available: \url{https://arxiv.org/abs/2304.13705}
\BIBentrySTDinterwordspacing

\bibitem{fang2023anygrasprobustefficientgrasp}
\BIBentryALTinterwordspacing
H.-S. Fang, C.~Wang, H.~Fang, M.~Gou, J.~Liu, H.~Yan, W.~Liu, Y.~Xie, and C.~Lu, ``Anygrasp: Robust and efficient grasp perception in spatial and temporal domains,'' 2023. [Online]. Available: \url{https://arxiv.org/abs/2212.08333}
\BIBentrySTDinterwordspacing

\end{thebibliography}

\end{document}